\documentclass[runningheads]{llncs}
\usepackage[T1]{fontenc}
\usepackage{graphicx,verbatim}
\usepackage{booktabs}
\usepackage{multirow}
\usepackage{placeins}
\usepackage{amssymb}
\usepackage{amsmath}
\usepackage{hyperref}
\usepackage{marvosym}
\begin{document}
\title{Calibrated Confidence Expression for Radiology Report Generation}
%
\author{David Bani-Harouni\inst{1,2}\thanks{Equal Contribution} \and
Chantal Pellegrini\inst{1,2}\textsuperscript{$\star$} \and
Julian Lüers\inst{1} \and
Su Hwan Kim\inst{3,4} \and
Markus Baalmann\inst{5} \and
Benedikt Wiestler\inst{2,4,6} \and
Rickmer Braren\inst{3,5} \and
Nassir Navab\inst{1,2} \and
Matthias Keicher\inst{1,2}
}


%
\authorrunning{D. Bani-Harouni, C. Pellegrini et al.}
%
\institute{Computer Aided Medical Procedures, Technical University of Munich, Germany \email{\{david.bani-harouni,chantal.pellegrini\}@tum.de} \and
Munich Center for Machine Learning (MCML), Germany \and
Department of Diagnostic and Interventional Radiology, TUM Klinikum rechts der Isar, Germany \and
Department of Diagnostic and Interventional Neuroradiology, TUM Klinikum rechts der Isar, Germany \and
Department of Diagnostic and Interventional Radiology and Nuclear Medicine, University Medical Center Hamburg-Eppendorf, Germany \and
AI for Image-Guided Diagnosis and Therapy, Technical University of Munich
}

\maketitle              
\begin{abstract}
Safe deployment of Large Vision-Language Models (LVLMs) in radiology report generation requires not only accurate predictions but also clinically interpretable indicators of when outputs should be thoroughly reviewed, enabling selective radiologist verification and reducing the risk of hallucinated findings influencing clinical decisions. One intuitive approach to this is verbalized confidence, where the model explicitly states its certainty. However, current state-of-the-art language models are often overconfident, and research on calibration in multimodal settings such as radiology report generation is limited. To address this gap, we introduce ConRad (Confidence Calibration for Radiology Reports), a reinforcement learning framework for fine-tuning medical LVLMs to produce calibrated verbalized confidence estimates alongside radiology reports. We study two settings: a single report-level confidence score and a sentence-level variant assigning a confidence to each claim. Both are trained using the GRPO algorithm with reward functions based on the logarithmic scoring rule, which incentivizes truthful self-assessment by penalizing miscalibration and guarantees optimal calibration under reward maximization. Experimentally, ConRad substantially improves calibration and outperforms competing methods. In a clinical evaluation we show that ConRad's report level scores are well aligned with clinicians' judgment. By highlighting full reports or low-confidence statements for targeted review, ConRad can support safer clinical integration of AI-assistance for report generation. Our code is available at \href{https://github.com/ChantalMP/Conrad}{\texttt{https://github.com/ChantalMP/Conrad}}.


\end{abstract}
\section{Introduction}
Recent progress in AI for radiology shows growing potential to assist with rising clinical demands \cite{kaur2022methods,acosta2024impact}. Specifically for report generation, Large Vision–Language Models (LVLMs) have improved generation of coherent and correct radiology reports directly from medical images \cite{chexagent-2024,bannur2024maira,pellegrini2025radialog,lee2025cxr,sellergren2025medgemma}. However, outputs generated by LVLMs are prone to issues such as hallucinations, while often being overly confident in their prediction \cite{hadi2023large,xu2025delusions}. Therefore, for such systems to be clinically useful, they must provide not only accurate outputs but also trustworthy, clinically actionable indicators of output quality, signaling when clinicians can rely on model findings and when in-depth verification is needed. One promising strategy is letting the model predict not only an output but also a confidence estimate, an indicator of how much the output should be trusted. If confidence is well-calibrated, it can support report-level triage and prioritization for selective human review as well as targeted verification of report passages, thereby reducing the effort for oversight without compromising safety.\looseness=-1

Confidence calibration has been widely studied for large language models \cite{wang2023calibration,geng2024survey}, spanning post-hoc black/white-box estimators and trained external predictors \cite{azaria2023internal}. Verbalization of confidence has been targeted using supervised training on pseudo-ground-truth \cite{zhang2024r,kang2025unfamiliar} or reinforcement learning against human feedback \cite{leng2024taming} or correctness measures \cite{bani2026rewarding,zhang2025reinforcement}. In radiology report generation, uncertainty has been explored as clinical uncertainty phrasing learned from report language \cite{wangcurv,yan2024diagnose}, however this does not target the model's inherent uncertainty but aims to reproduce the confidence expressions of radiologists. Other methods model uncertainty estimation via stochastic sampling or semantic consistency across multiple generated reports \cite{wang2024trust,wang2025semantic}, or via post-hoc auditing that validates high-level extracted findings against external image classifiers \cite{warr2024quality}. These approaches are, however, inefficient at inference as they require multiple generations to estimate uncertainty. They further do not train explicitly for improving the internal confidence calibration of the model and do not enable it to express calibrated confidences alongside reports.

Inspired by the success of reinforcement learning with proper scoring rules for calibrating confidence in large language models \cite{bani2026rewarding,zhang2025reinforcement}, we propose ConRad, which extends scoring-rule–based confidence calibration to multimodal, long-form radiology report generation. ConRad fine-tunes medical LVLMs to express verbalized confidence jointly with the report text, and supports both report-level and fine-grained sentence-level confidence estimates to guide targeted verification. We combine Group Relative Policy Optimization (GRPO) \cite{shao2024deepseekmath} with a logarithmic scoring-rule reward and a continuous correctness signal provided by an automated report-quality metric, enabling calibration without manual confidence supervision. In summary, we propose: 1) a novel RL-based approach for verbalized confidence calibration in multimodal long-form report generation; 2) two settings for report- and sentence-level confidence expression enabling confidence-based verification; and 3) a reward design that couples policy optimization of the proper scoring rule with a clinical report quality metric. We evaluate ConRad on MIMIC-CXR, demonstrating substantial calibration gains over strong confidence estimation baselines, supporting safer integration of LVLM report generators into clinical workflows via confidence-guided review.

\section{Methodology}
We propose ConRad, a method for calibrating verbalized confidence in medical Large Vision-Language Models (LVLMs) during radiology report generation, enabling confidence-guided review in clinical workflows. ConRad supports both report- and sentence-level confidence scores, allowing clinicians to quickly identify which reports or statements need targeted verification. Our method fine-tunes LVLMs to jointly generate findings and numerical confidence scores by formulating the confidence expression as a Markov Decision Process, employing Reinforcement Learning (RL) to align the model's verbalized confidence with an external correctness assessment. In this setting, confidence token predictions are the actions, while prompt and report represent the state. 

\begin{figure}[tb]
    \centering
   \includegraphics[width=\linewidth]{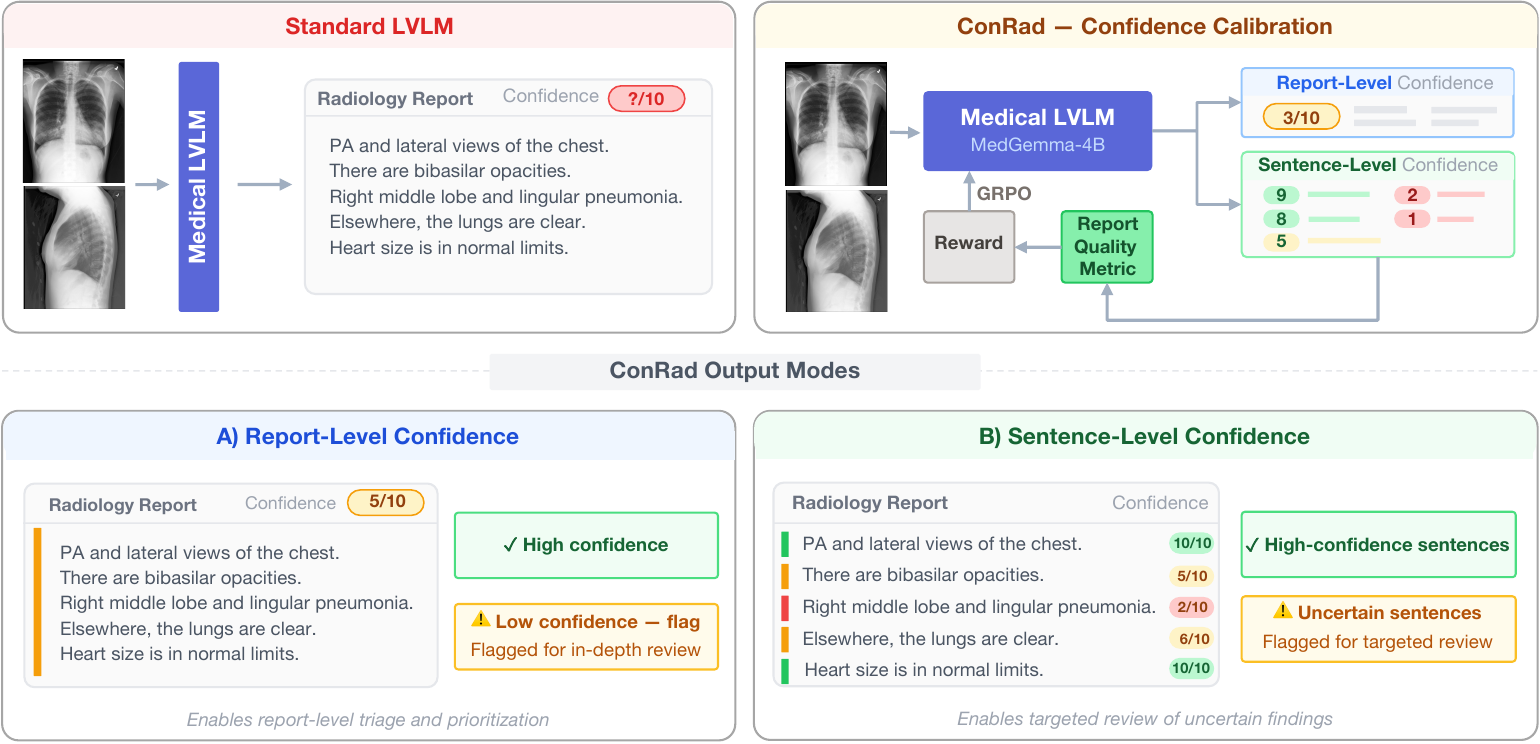}
    \caption{Overview of ConRad. While standard medical LVLMs provide reports without any confidence measure, fine-tuning them with ConRad enables calibrated confidence expression on a report- and sentence-level that could be used for targeted review.}
    \label{fig:graphical_abstract}
\end{figure}

\subsection{Problem Formulation: Confidence and Calibration}
Let $v = (v_1, \dots, v_k)$ denote a set of input medical images of a radiological study. The model generates a textual report $r$ and a verbalized scalar confidence score, which we scale to $\hat{p} \in [0, 1]$. We define confidence as the model's self-reported quality estimate of the report $r$ according to a ground-truth correctness metric $s \in [0, 1]$ (e.g., semantic similarity to a ground-truth reference report).
In this setting perfect calibration is achieved, when the verbalized confidence matches the expected correctness:
\begin{equation}
    \mathbb{E}[s \mid \hat{p} = x] = x, \quad \forall x \in [0,1].
\end{equation}

\subsection{Confidence Generation}
We employ a pre-trained LVLM that accepts images $v$ and prompt $l$ to generate a response. We investigate two granularity levels for confidence expression:\\
\textbf{Report-level Scenario:} The model generates a complete report $r$ followed by a single confidence score $\hat{p}$ representing the confidence of the entire report.\\
\textbf{Sentence-level Scenario:} A report is generated as a sequence of $m$ sentences, $r = (r_1, \dots, r_m)$. After each sentence $r_i$, the model generates a confidence score $\hat{p}_i$ to form an output $((r_1, \hat{p}_1), \dots, (r_m, \hat{p}_m))$.

\subsection{Reinforcement Learning for Calibration}
We train confidence calibration using reinforcement learning. Unlike supervised LLM fine-tuning on expressing correctness values, which treats numerical confidence tokens as independent categorical labels and can ignore their ordinal structure, our RL approach utilizes a continuous reward signal that varies continuously with the distance of predicted confidence and correctness, so near-misses are preferred over large miscalibration. This incentivizes the model to learn a gradual internal representation of confidence as also shown by Zhang et al. \cite{zhang2025reinforcement}.

To incentivize calibration, the reward function $R$ must be maximized when the predicted confidence $\hat{p}$ aligns with the true correctness probability. We base our rewards on the Logarithmic Scoring Rule, a strictly proper scoring rule where the expected reward is uniquely maximized if and only if the predicted probability matches the true distribution. Note, that our goal is not to improve the quality of the report generation capabilities itself but only the confidence calibration of the model. We therefore only calculate a loss for the tokens involved in confidence expression, ensuring that the task performance remains stable. \\
\textbf{Report-level Reward ($R_\text{R}$).}
In the report-level scenario, the reward is computed with respect to a continuous correctness score $s \in [0, 1]$ evaluating the entire report. We define the reward as the logarithmic scoring rule with $s$ as the target and $\lambda$ as scaling factor:
\begin{equation}
    R_\text{R}(r, \hat{p}, s) = \lambda \cdot \left[ s \log(\hat{p}) + (1 - s) \log(1 - \hat{p}) \right]
\end{equation}
This formulation can be thought of as policy optimization of the cross-entropy between predicted confidence and correctness target that penalizes deviations between the predicted $\hat{p}$ and the measured correctness $s$.\\
\textbf{Sentence-level Reward ($R_\text{S}$).}
In the sentence-level scenario, correctness is evaluated with a binary correctness score $f_i \in \{0, 1\}$ for each sentence $i$. Since $f_i$ is binary, we define the reward as the mean logarithmic score over all sentences:
\begin{equation}
    R_\text{S}(r, \hat{p}, f) = \frac{\lambda}{m} \sum_{i=1}^{m} \left[ f_i \log(\hat{p}_i) + (1 - f_i) \log(1 - \hat{p}_i) \right]
\end{equation}
To ensure numerical stability, we clip confidence predictions to $[\varepsilon, 1-\varepsilon]$ with $\varepsilon$ sufficiently small.\\
\\
This reward structure encourages the model to be confident only when it is correct (high $s$ or $f=1$) and to express uncertainty (low $\hat{p}$) when it is likely to be incorrect, effectively calibrating the model via RL.

\section{Experiments}
\begin{table}[tb]
\centering
\caption{Report-level evaluation results. ECE and Pearson correlation serve as the primary evaluation metrics, GREEN score is included as a control to verify that report generation quality remains stable. 95 \% confidence intervals in brackets.}
\label{tab:single_conf_metrics}
\begin{tabular}{lccc}
\toprule
& \multicolumn{2}{c}{\textbf{Calibration Metrics}} \\
\cmidrule{2-3}
\textbf{Method} & \textbf{ECE} ($\downarrow$) & \textbf{Correlation} ($\uparrow$) & \textbf{GREEN} \\
\midrule
MedGemma Base \cite{sellergren2025medgemma} & - & - & 0.283 {\scriptsize [0.271, 0.296]}\\
Sequence Probability \cite{huang2025look} & 0.531 {\scriptsize [0.519, 0.542]} & 0.340 {\scriptsize [0.282, 0.388]} & 0.286 {\scriptsize [0.273, 0.299]}\\
P(True) \cite{kadavath2022language} & 0.399 {\scriptsize [0.380, 0.418]} & 0.166 {\scriptsize [0.113, 0.222]} & 0.287 {\scriptsize [0.275, 0.300]}\\
Self-Consistency \cite{lin2024generating} & 0.436 {\scriptsize [0.424, 0.448]} & 0.333 {\scriptsize [0.277, 0.393]} & 0.297 {\scriptsize [0.284, 0.310]}\\
Trained Probe \cite{azaria2023internal} & 0.121 {\scriptsize [0.110, 0.135]} & 0.372 {\scriptsize [0.318, 0.422]} & 0.299 {\scriptsize [0.287, 0.312]}\\
\hline
Verbalize Base & 0.642 {\scriptsize [0.630, 0.654]} & 0.136 {\scriptsize [0.078, 0.194]} & 0.257 {\scriptsize [0.245, 0.268]}\\
Verbalize Supervised & 0.206 {\scriptsize [0.189, 0.223]} & 0.160 {\scriptsize [0.089, 0.222]} & 0.252 {\scriptsize [0.240, 0.264]}\\
ConRad & \textbf{0.106} {\scriptsize [0.093, 0.118]} & \textbf{0.431} {\scriptsize [0.372, 0.487]} & 0.252 {\scriptsize [0.240, 0.264]}\\
\bottomrule
\end{tabular}
\end{table}

\noindent\textbf{Experimental Setup and Metrics.}
All main experiments use MIMIC-CXR \cite{johnson2019mimic}. Given all chest x-rays of a study and the Indication section, the model generates the Findings section and confidence estimates. Confidence is an integer in $\{0,\dots,10\}$, normalized to $[0,1]$ for evaluation. Calibration is measured with Expected Calibration Error (ECE); we additionally report Pearson correlation for report-level confidence, comparing to the continuous correctness score
and AUROC for sentence-level confidence, comparing to binary per-sentence correctness. 
To test zero-shot transfer, we evaluate on the IU-Xray \cite{demner2016preparing} dataset, using the test split defined in Rad-ReStruct \cite{pellegrini2023rad}.

We compare against representative black and white-box confidence estimation baselines: \textit{Verbalize Base} (zero-shot confidence via prompting), \textit{Verbalize Supervised} (SFT on explicit GREEN scores appended to the report text using cross-entropy loss), \textit{Sequence Probability} (mean token probability \cite{huang2025look}), \textit{P(True)} (two-stage self-assessment where the model predicts true or false after the report; confidence estimated from probabilities of true vs. false token \cite{kadavath2022language}), \textit{Self-Consistency} (confidence as agreement across multiple sampled reports \cite{lin2024generating}), and \textit{Trained Probe} (MLP predicts correctness from hidden states of the LLM \cite{azaria2023internal}).

\noindent\textbf{Implementation Details.}
Report-level confidence is trained on 3000 samples for one epoch; sentence-level confidence on 1500 samples for one epoch, both until convergence. We use a learning rate of $10^{-5}$ on an NVIDIA A40 GPU. We use a 4-bit quantized \texttt{medgemma-4b-it} \cite{sellergren2025medgemma} as report generation model and fine-tune only the language model component using LoRA \cite{hu2022lora} and GRPO \cite{shao2024deepseekmath}. GRPO is an RL algorithm for LLM finetuning. It samples multiple candidate outputs for the same input, normalizes and compares their rewards, and updates the model to increase the likelihood of the comparatively higher-reward outputs. As a correctness signal $s$, we use GREEN \cite{ostmeier2024green}, which is a common report generation metric that scores a generated report against the reference report in $[0,1]$ using an LLM-as-a-judge, evaluating the ratio of matched findings and clinically significant errors. For sentence-level correctness scoring, we utilize an adapted GREEN (Precision GREEN) $f$ that only evaluates whether the finding in that sentence is supported by the reference report. During training, the loss is masked to the confidence tokens, constraining optimization to confidence prediction rather than report content. We set the reward scaling $\lambda$ to 100. Outputs that violate the required format receive a fixed negative reward. For the trained probe, we also train on 3000 samples for 10 epochs with early stopping on the validation set with a learning rate of $10^{-4}$ and use the hidden states of layer 35.

\begin{figure*}[tb]
    \centering
    \setlength{\tabcolsep}{3pt} 
    \renewcommand{\arraystretch}{1.0}

    \begin{tabular}{ccc}
        \;\;\quad\small\textbf{Report-level} &
        \;\;\quad\small\textbf{Sentence-level} &
        \;\;\quad\small\textbf{Report-level OOD} \\

        \includegraphics[width=0.31\textwidth]{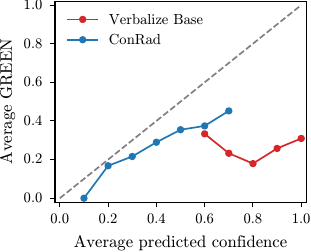} &
        \includegraphics[width=0.31\textwidth]{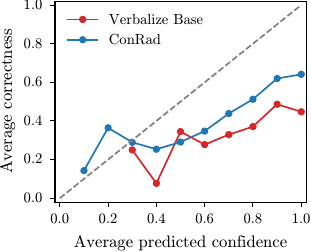} &
        \includegraphics[width=0.31\textwidth]{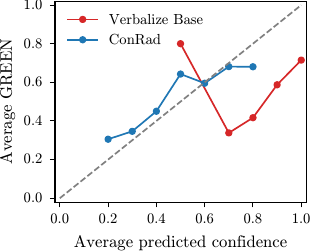} \\[-0.2em]

        \includegraphics[width=0.31\textwidth]{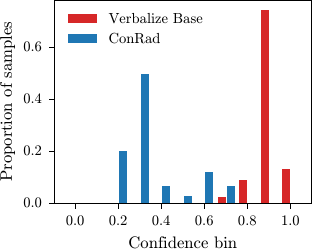} &
        \includegraphics[width=0.31\textwidth]{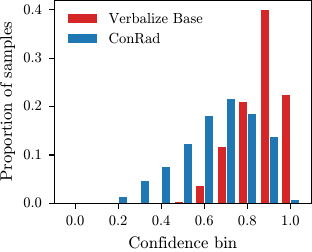} &
        \includegraphics[width=0.31\textwidth]{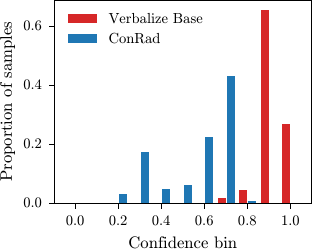} \\
    \end{tabular}

    \caption{Verbalize Base vs.\ ConRad: calibration curves (top row) and confidence distributions (bottom row) for report-level, sentence-level, and out-of-distribution (OOD).
    }
    \label{fig:calibration_illustrations_3x2}
\end{figure*}

\section{Results and Discussion}
\subsubsection{Report-level Confidence Calibration.}
Table~\ref{tab:single_conf_metrics} summarizes the performance in the report-level confidence scenario. ConRad outperforms all baselines, reducing the ECE by over 80\% compared to the base model and achieving the highest Pearson correlation with factual correctness. As shown in Figure~\ref{fig:calibration_illustrations_3x2}, the fine-tuned model shifts from the base model's extreme overconfidence to a more balanced distribution and a calibration curve closer to the ideal diagonal. We observe a decrease in GREEN score when prompting MedGemma to verbalize its confidence (Verbalize Base) compared to report-only prompting, likely caused by the shift from the MedGemma training setting. At the same time the report correctness remains stable compared to Verbalize Base during training with ConRad. This is desired as we do not want the training to influence the report generation process itself.\looseness=-1

While the Trained Probe achieves a competitive ECE (Table~\ref{tab:single_conf_metrics}), it depends on internal hidden states, whereas ConRad offers a solution that verbalizes confidence directly. Furthermore, unlike supervised fine-tuning which treats tokens as independent categories, our RL approach preserves the ordinal structure of confidence. This is consistent with the softmax distribution over the confidence-token vocabulary, where ConRad assigns the highest probability to the predicted confidence level and progressively smaller probabilities to adjacent levels, resulting in a smooth notion of “nearby” certainty. In contrast, supervised fine-tuning as in Verbalize Supervised disrupts this ordering, producing a less coherent probability pattern across tokens that suggests the model is learning a set of unrelated labels rather than a graded confidence scale.\\
\noindent\textbf{Sentence-level Confidence Calibration.}
We additionally explore a sentence-level confidence scenario. Here, ConRad evaluates each sentence confidence separately against a binary correctness target. As shown in Table~\ref{tab:multi_conf_metrics}, our method reduces sentence-level ECE by $\sim$40\% and improves AUROC compared to the base model. Figure~\ref{fig:calibration_illustrations_3x2} confirms this improvement, showing a calibration curve that tracks the diagonal more closely than the base model. Also in this setting, the GREEN score remains stable after training with ConRad. This experiment also highlights an advantage of our RL approach over SFT for calibration in settings with binary targets. Through its RL-based formulation, the model trained with ConRad is still able to learn fine-grained confidence predictions, while Verbalize Supervised can only be trained to reproduce these binary targets leading to suboptimal calibration. Overall, sentence-level calibration remains more challenging than report-level calibration, where we observe stronger overall results.\looseness = -1
\begin{table}[tb]
\centering
\caption{Sentence-level evaluation results. ECE and AUROC assess calibration and discriminative power, GREEN score is included as a control to verify that report generation quality remains stable. 95 \% confidence intervals in brackets.}
\label{tab:multi_conf_metrics}
\begin{tabular}{lccc}
\toprule
& \multicolumn{2}{c}{\textbf{Calibration Metrics}} \\
\cmidrule{2-3}
\textbf{Method} & \textbf{ECE} ($\downarrow$) & \textbf{AUROC} ($\uparrow$) & \textbf{GREEN} \\
\midrule
MedGemma Base \cite{sellergren2025medgemma} & - & - & 0.283 {\scriptsize [0.271, 0.296]}\\
Verbalize Base & 0.438 {\scriptsize [0.423, 0.453]} & 0.558 {\scriptsize [0.543, 0.572]} & 0.268 {\scriptsize [0.257, 0.279]}\\
Verbalize Supervised & 0.446 {\scriptsize [0.433, 0.460]} & 0.552 {\scriptsize [0.538, 0.565]} & 0.267 {\scriptsize [0.257, 0.280]}\\
ConRad & \textbf{0.239} {\scriptsize [0.225, 0.253]} & \textbf{0.633} {\scriptsize [0.620, 0.647]} & 0.262 {\scriptsize [0.250, 0.274]}\\
\bottomrule
\end{tabular}
\end{table}
\\
\noindent\textbf{Confidence-Based Flagging.}
\label{subsec:filtering}
ConRad enables flagging sentences or reports with low confidence, making clinicians aware of findings that have an increased probability of being wrong.
To assess this use case practically, we filter sentences based on their predicted confidence $\hat{p}$ and evaluate the remaining sentences with Precision GREEN. Table~\ref{tab:ordinal_confidences} demonstrates that the factual precision increases monotonically with the confidence threshold. 
\begin{table}[tb]
\centering
\caption{Number of sentences remaining after filtering by confidence threshold and Precision GREEN on the filtered sentences.}
\label{tab:ordinal_confidences}
\setlength{\tabcolsep}{3pt}
\begin{tabular}{lcc|cc}
\toprule
\multirow{2}{*}{\textbf{Threshold}} & \multicolumn{2}{c|}{\textbf{Verbalize Base}} & \multicolumn{2}{c}{\textbf{ConRad}} \\
 & \shortstack{\#Sentences} & \shortstack{Precision GREEN} & \shortstack{\#Sentences} & \shortstack{Precision GREEN} \\
\midrule
All                  & 6131 & 0.425 & 6317 & 0.421 \\
Confidence $\geq$ 6  & 6081 & 0.426 & 4634 & 0.471 \\
Confidence $\geq$ 8  & 5128 & 0.447 & 2106 & 0.560 \\
Confidence = 10      & 1379 & 0.447 & 53 & 0.642 \\
\bottomrule
\end{tabular}
\end{table}
However, this comes with an expected trade-off between precision and completeness. The standard GREEN score over the remaining report decreases (0.262 (All) → 0.097 (confidence = 10)), indicating omission of uncertain but potentially correct content. This suggests that low-confidence sentences should be flagged for human review rather than automatically removed.
In contrast, filtering based on Verbalize Base confidences yields only marginal Precision GREEN gains, despite a similar drop in GREEN score (0.268 → 0.121). This indicates that the base model’s expressed confidence is poorly aligned with factual correctness, whereas ConRad’s confidence provides actionable signals for targeted review.\\
\noindent\textbf{Performance on OOD Data.}
To test generalization of ConRad beyond the training data, we evaluate on IU-Xray in a strict zero-shot setting, where ConRad is trained on MIMIC-CXR only, with no IU-Xray exposure. ConRad's calibration generalizes substantially better than the base model's with ECE improving $\sim$9$\times$ from 0.310 to 0.034, while maintaining comparable report quality. As seen in Figure \ref{fig:calibration_illustrations_3x2}, Verbalize Base is severely overconfident, while a much broader range of confidence values is used when applying a model fine-tuned with ConRad.\\
\noindent\textbf{Clinical Evaluation.}
To assess the alignment of report confidence to radiologist judgment, we conduct a small-scale clinical evaluation on 50 reports with three raters. Each rater scores every sentence on a 5-point Likert scale from \textit{Reject} to \textit{Accept}. We derive two report-level clinician ratings from these sentence annotations. \textit{Mean aggregation} averages scores across all sentences and raters, yielding a continuous report-level quality estimate. \textit{All accepted} defines a binary acceptability criterion: a report is marked acceptable only if all sentences receive a score of at least 4 from a majority of raters. We report correlation, AUROC and ECE with respect to clinician ratings and model confidences. Table~\ref{tab:clinical_eval} shows that ConRad achieves stronger alignment with clinician judgment across both aggregation settings than Verbalize Base, indicating that the calibration gains of ConRad transfer to expert evaluation, supporting its potential for report-level triage.

\begin{table}[tb]
\centering
\caption{Report-level clinical evaluation on 50 reports with 3 raters. Two report-level targets are derived from sentence-wise Likert ratings: \emph{Mean aggregation} and \emph{All accepted}. We report Spearman correlation, AUROC and ECE.}
\label{tab:clinical_eval}
\small
\setlength{\tabcolsep}{5pt}
\begin{tabular}{l l c c c}
\hline
Aggregation & Method & Correlation & AUROC & ECE \\
\hline
\multirow{2}{*}{Mean aggregation}
& Verbalize Base & 0.069  & 0.519 & 0.537 \\
& ConRad & \textbf{0.268} & \textbf{0.644} & \textbf{0.099} \\
\hline
\multirow{2}{*}{All accepted}
& Verbalize Base & -0.036 & 0.489 & 0.842 \\
& ConRad & \textbf{0.103} & \textbf{0.610} & \textbf{0.220} \\
\hline
\end{tabular}
\end{table}

\section{Conclusion}
In this work, we introduced ConRad, a reinforcement-learning framework that fine-tunes medical LVLMs to jointly generate radiology reports and verbalized, calibrated confidence. ConRad supports both a global report-level, and a local sentence-level confidence expression. Our results show that confidence calibration is improved compared to strong baselines, while keeping the report generation performance unchanged. Such calibrated confidence expression, integrated in the report generation process, can enable more efficient expert verification, guided by the model's confidence, opening a path towards safe integration of AI support tools into clinical practice. 

\begin{credits}
\subsubsection{\ackname} 
The authors gratefully acknowledge the financial support by the Bavarian Ministry of Economic Affairs, Regional Development and Energy (StMWi) under project ThoraXAI (DIK-2302-0002), and the German Research Foundation (DFG, grant 469106425 - NA 620/51-1).

\subsubsection{\discintname}
The authors have no competing interests to declare that are relevant to the content of this article. 
\end{credits}

%
%
%

\newpage

\bibliographystyle{splncs04}
\bibliography{paper-2409}

\end{document}